\documentclass[10pt]{iopart}

\expandafter\let\csname equation*\endcsname\relax
\expandafter\let\csname endequation*\endcsname\relax
\usepackage{amssymb}
\usepackage{amsmath}
\usepackage{natbib}

\usepackage{xcolor}
\usepackage{subfigure}
\usepackage{ctable}
\usepackage{makecell}
\usepackage{caption} 
\usepackage[shortlabels]{enumitem}
\usepackage{hyperref}
\usepackage{xspace}
\usepackage{lineno}

\usepackage{makecell}
\usepackage{multirow}
\usepackage{ctable}
\usepackage{ulem} 
\usepackage{bm}

\usepackage{xcolor}

\begin{document}

\title[DeepAdversaries]{DeepAdversaries: Examining the Robustness of Deep Learning Models for Galaxy Morphology Classification}

\author{Aleksandra \'Ciprijanovi\'c${}^1$, Diana Kafkes${}^1$, Gregory Snyder${}^2$, F. Javier S\'{a}nchez${}^1$, Gabriel Nathan Perdue${}^1$, Kevin Pedro${}^1$, Brian Nord${}^{1,3,4}$, Sandeep Madireddy${}^5$, Stefan M.\ Wild${}^5$}

\address{${}^1$Fermi National Accelerator Laboratory, Batavia, IL 60510, USA}
\address{${}^2$Space Telescope Science Institute, Baltimore, MD 21218, USA}
\address{${}^3$Department of Astronomy and Astrophysics, University of Chicago, Chicago, IL 60637, USA}
\address{${}^4$Kavli Institute for Cosmological Physics, University of Chicago, Chicago, IL 60637, USA}
\address{${}^5$Mathematics and Computer Science Division, Argonne National Laboratory, Lemont, IL 60439, USA}

\ead{aleksand@fnal.gov}

\begin{abstract}
With increased adoption of supervised deep learning methods for work with cosmological survey data, the assessment of data perturbation effects (that can naturally occur in the data processing and analysis pipelines) and the development of methods that increase model robustness are increasingly important.
In the context of morphological classification of galaxies, we study the effects of perturbations in imaging data.
In particular, we examine the consequences of using neural networks when training on baseline data and testing on perturbed data.

We consider perturbations associated with two primary sources: 1) increased observational noise as represented by higher levels of Poisson noise and 2) data processing noise incurred by steps such as image compression or telescope errors as represented by one-pixel adversarial attacks.
We also test the efficacy of \textit{domain adaptation} techniques in mitigating the perturbation-driven errors.
We use classification accuracy, latent space visualizations, and latent space distance to assess model robustness in the face of these perturbations.
For deep learning models without domain adaptation, we find that processing pixel-level errors easily flip the classification into an incorrect class and that higher observational noise makes the model trained on low-noise data unable to classify galaxy morphologies. 
On the other hand, we show that training with domain adaptation improves model robustness and mitigates the effects of these perturbations, improving the classification accuracy up to 23\% on data with higher observational noise. 
Domain adaptation also increases up to a factor of ${\approx}2.3$ the latent space distance between the baseline and the incorrectly classified one-pixel perturbed image, making the model more robust to inadvertent perturbations. 
Successful  development  and  implementation  of methods that increase model robustness in  astronomical survey pipelines will help pave the way for many more uses of deep learning for astronomy.

\end{abstract}

%
\noindent{\it Keywords}: convolutional neural networks, deep learning, model robustness, adversarial attacks, galaxy morphological classification, sky surveys 
%
%

\ioptwocol
%

\section{Introduction}
\label{sec: inroduction}

The success of deep learning models across a broad range of science applications is in part driven by their inherent flexibility. For example, deep neural networks can be trained to use features that represent a wide variety of patterns in the data. 
However, the features these neural networks contain are often incomprehensible by humans, and the models they produce can be brittle, especially when applied outside the intended circumstances.
One such change in circumstances happens when trained models are applied to data that contain perturbations, which can be intentional or accidental in origin.
The effective use of deep learning tools requires a detailed exploration and accounting of failure modes amidst possible data perturbations.

Adversarial attacks are inputs specifically crafted to confuse susceptible neural networks~\citep{SZ2013,YH2019}.
Often, adversarial examples are thought to arise from non-robust features that can easily be learned by overly-parameterized models~\citep{IS2019}.
Some attacks rely on access to network information, such as network architecture, trained weights, and internal gradients~\citep{SZ2013,GS2014}.
One of the most well-known examples of this type of attack is a correctly classified image of a panda that is flipped to the class ``gibbon'', with very high probability, after the addition of imperceptible but well-crafted noise, produced by the ``fast gradient sign method''~\citep{GS2014}. 
Alternatively, black-box attacks do not require information about the trained model~\citep{BH2017,CZ2017}. For example, analysis of the widely used benchmark datasets CIFAR-10~\citep{CIFAR10} and ImageNet~\citep{IMG2009} shows that ${\approx}68\%$ and ${\approx}16\%$ of images, respectively, can be flipped to an incorrect class by changing or ``attacking'' just one pixel of the image~\citep{SV2019}.
Furthermore, perturbations of real-world objects can cause significant problems; for example, well-placed stickers on traffic signs have caused autonomous vehicles to misclassify stop signs~\citep{EE2017}.

Beyond the extreme of adversarial attacks, readily occurring or accidental data perturbations---including image compression, blurring (via the point spread function), and the addition of observational (often simple Gaussian or Poisson) noise, instrument  readout  errors, dead camera pixels---can significantly degrade or imperil model performance~\citep{GD2016, DK2016,DK2017,FG2019} in astronomy applications. In the sciences, adversarial attacks can be used as a proxy for some of these naturally occurring perturbations to obtain a deeper understanding of model performance and robustness. This is crucial for successful implementation of deep learning in astronomy experiments, in particular for real-time data acquisition and processing.

Deep learning is used with increasing frequency for a variety of tasks in cosmology, from science analysis to data processing.
For example, convolutional neural networks (CNNs) and more complex residual neural networks have been used to classify/identify a variety of objects and patterns, such as:  low surface brightness galaxies~\citep{TC2021}, merging galaxies~\citep{CS2020}, galaxy-galaxy strong lenses~\citep{LM2018} or Sunyaev-Zel'dovich galaxy clusters~\citep{LH2021}.
Furthermore, deep learning has often been used for regression tasks such as measuring galaxy properties from $21\,\mathrm{cm}$ maps~\citep{PM2022}, or constraining cosmological parameters from weak lensing maps~\citep{FK2019}.
Finally, deep learning can also be used to automate multiple tasks in large astronomical surveys, including telescope survey scheduling~\citep{NY2019,AH2019}, cleaning astronomical data sets of ghosts and scattered-light artifacts~\citep{TC2021b}, image denoising~\citep{GV2022}, and data processing and storing~\citep{LW2021}.
The use of deep learning is likely to grow commensurately with the size and complexity of modern and next-generation cosmic surveys, such as the Dark Energy Survey~\citep[DES;][]{DES2016}, the Hyper Suprime-Cam Subaru Strategic Program~\citep[HSC-SSP;][]{HSC2018}, the Rubin Observatory Legacy Survey of Space and Time~\citep[LSST;][]{IK2019}, Euclid\footnote[1]{https://www.cosmos.esa.int/web/euclid}, the Nancy Grace Roman Space Telescope\footnote{https://roman.gsfc.nasa.gov}, the Subaru Prime Focus Spectrograph~\citep[PSF;][]{PFS2015}, and the Dark Energy Spectroscopic Instrument~\citep[DESI;][]{DESI2016,DESI2016b}. 

Most approaches to defend from adversarial attacks~\citep{HD2019} can be divided into: 1) reactive measures, which focus on detecting the attack after the model is built~\citep{LI2017,MG2017,FC2017}, cleaning the attacked image~\citep{GR2014}, or verifying the network properties~\citep{KB2017}; and 2) proactive measures, which aim to increase model robustness before adversarial attacks are produced~\citep{YH2019}. 
In the sciences, where adversarial attacks are not targeted but can accrue as a natural part of the data acquisition and storage process, the second group of the defense strategies is more relevant.
Some of the methods in this group include network distillation~\citep{PM2016}, adversarial (re)training~\citep{GS2014,madry2018towards,deng2020interpreting}, and probabilistic modeling to provide uncertainty quantification~\citep{BM2017,AG2017,wicker2021bayesian}. 
More recently, it has also been shown that viewing a neural architecture as a dynamical system (referred to as implicit neural networks) and incorporating higher-order numerical schemes can improve robustness to adversarial attacks~\citep{li2020implicit}. 

Domain adaptation~\citep{C2017,WD2018,GD2020} comprises another group of methods that could prove useful to increase the robustness of deep learning models against these naturally occurring image perturbations. These techniques are useful when training models that need to perform well on multiple datasets at the same time. 
In contrast with the previously mentioned approaches, domain adaptation enables the model to learn domain-invariant features, which are present in multiple datasets and therefore more generalizable, thus improving models' robustness to inadvertent perturbations. 
Domain adaptation techniques can be categorized into: 1) distance-based methods such as Maximum Mean Discrepancy (MMD)~\citep{GB2007,GB2008}, Deep Correlation Alignment (CORAL)~\citep{SS2016}, Central Moment Discrepancy (CMD)~\citep{ZM20191}; and 2) adversarial-based methods such as Domain Adversarial Neural Networks (DANN)~\citep{GU2016} and Conditional Domain Adversarial Networks (CDAN)~\citep{LC2017}.

In the context of astronomical observations, the different domains may be simulated and observed data, or data from multiple telescopes. 
With domain adaptation, the model can be guided to ignore discrepancies across datasets, including different signal-to-noise levels, noise models, and PSFs. 
In~\cite{CS2020}, the authors show that a simple algorithm trained to distinguish merging and non-merging galaxies is rendered useless after the inclusion of observational noise. 
In~\cite{CK2020,CK2021} the authors study domain adaptation as a way to draw discrepant astronomical data distributions closer together, thereby increasing model robustness. 
Using domain adaptation, the authors were able to create a model trained on simulated images of merging galaxies from the Illustris-1 cosmological simulation~\citep{VG2014}, which also performs well on simulated data that includes observational noise. Furthermore, by using domain adaptation the authors were able to bridge the gap between simulated and observed data and create a model trained on simulated Illustris-1 data that performs well on the real Sloan Digital Sky Survey images ~\citep[SDSS;][]{LS2008,LS2010,DK2010}.

We posit this robustness is also directly applicable to combating inadvertent pixel-level perturbations coming from image compression or telescope errors. 
Domain adaptation methods are well suited for astronomy applications since they allow one to utilize previous observations or simulated data to increase the robustness of the model for new datasets. 
More importantly, domain adaptation methods can even be used when one of the datasets does not include labels. Several relevant cases include working with newly observed unlabeled data, which cannot directly be used to train a model, or fine tuning the weights (via transfer learning) of a model previously trained on old observations or simulations~\citep{TH2018,DS2019, TC2021}.

In this work, we use simulated data to explore the effects of inadvertent image perturbations that can arise in complex scientific data processing pipelines, including those that will be used by the Vera C.\ Rubin Observatory's LSST~\citep{IK2019}.
As the context for our tests, we use the problem of galaxy morphology classification (spiral, elliptical, and merging galaxies), using images and catalog data of galaxy morphology from the large-volume cosmological magneto-hydrodynamical simulation IllustrisTNG100~\citep{NS2019}.
We emulate LSST processing and observations in our images: our baseline dataset representing low-noise observations is generated by applying an exposure time equivalent to ten years of observing. 
For the first perturbation to the data, we explore the effects of larger observational noise by creating the high-noise observations, which correspond to one year of observing. 
We also explore pixel-level perturbations---representing effects such as data compression, instrument readout errors, cosmic rays, and dead camera pixels---which are produced through optimized one-pixel attacks~\citep{SV2017}.
We train our networks---a simple few-layer CNN we call \textit{ConvNet} and a more complex \textit{ResNet18}~\citep{HZ2016}---on baseline data and then test on noisy and one-pixel attacked data. During model training, we employ domain adaptation, to investigate the potential benefits of these methods for increasing model robustness to image perturbations, compared to regular training without domain adaptation.
Furthermore, we analyze the network latent spaces to assess the robustness of our models due to these data perturbations and training procedures.

In Section~\ref{sec: Data}, we describe the simulation and how we create our datasets, as well as details about the image perturbations we explore.
In Section~\ref{sec: Networks}, we describe the deep learning models we use, and in Section~\ref{sec: domain_adapt}, we introduce domain adaptation and how it is implemented in our experiments. 
In Section~\ref{sec: latent}, we introduce visualization methods that are used to explore the latent space of our models. 
We present our results in Section~\ref{sec: results}, with a discussion and conclusion in Section~\ref{sec: DandC}.

\section{Data}
\label{sec: Data}

\begin{sloppypar}When creating our dataset, we use IllustrisTNG100~\citep{MV2018,NP2018,SP2018,NP2018,PH2018,NS2019} -- a state-of-the-art cosmological magneto-hydrodynamical simulation that includes gas, stars, dark matter, supermassive black holes, and magnetic fields.
We extract galaxy images in ($g,r,i$) filters \footnote{We use database filter keys \texttt{psi\_g}, \texttt{psi\_r}, and \texttt{psi\_i}.} from snapshots at two redshifts: 95 ($z=0.05$) and 99 ($z=0$). 
Finally, we convert all data to an effective redshift of $z=0.05$, to create a larger single-redshift dataset.\end{sloppypar}

\begin{figure}
	\includegraphics[width=\columnwidth]{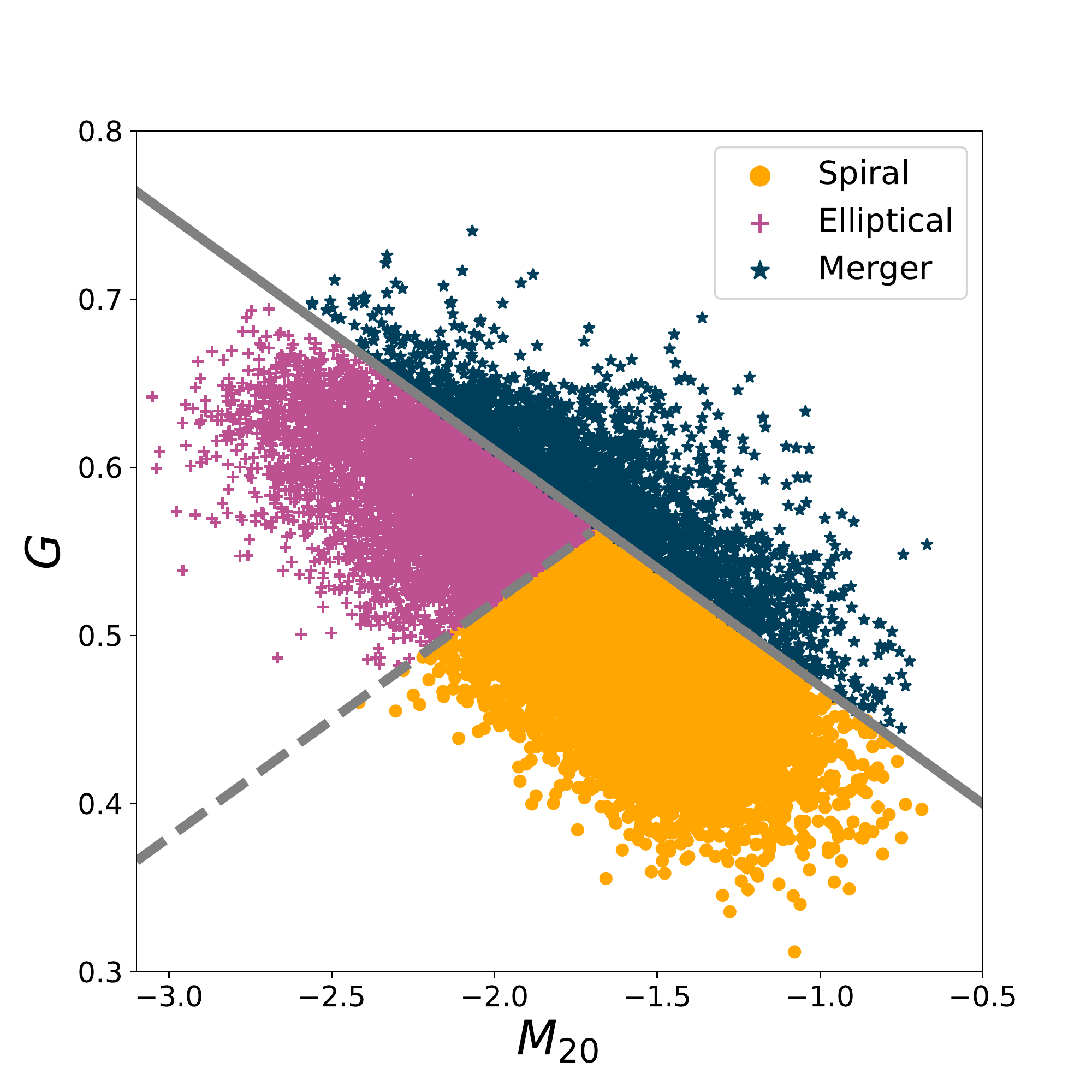}\\
    \caption{The $G$-$M_{20}$ ``bulge statistic'' plot for our entire dataset made from the two IllustrisTNG100 snapshots. 
    The solid gray line distinguishes between merging (dark blue stars) and non-merging systems. 
    The latter includes spiral (orange circles) and elliptical (violet plus signs) galaxies, which are separated by a dashed gray line.}
    \label{fig:classes}
\end{figure}

\begin{figure*}
	\includegraphics[width=\linewidth]{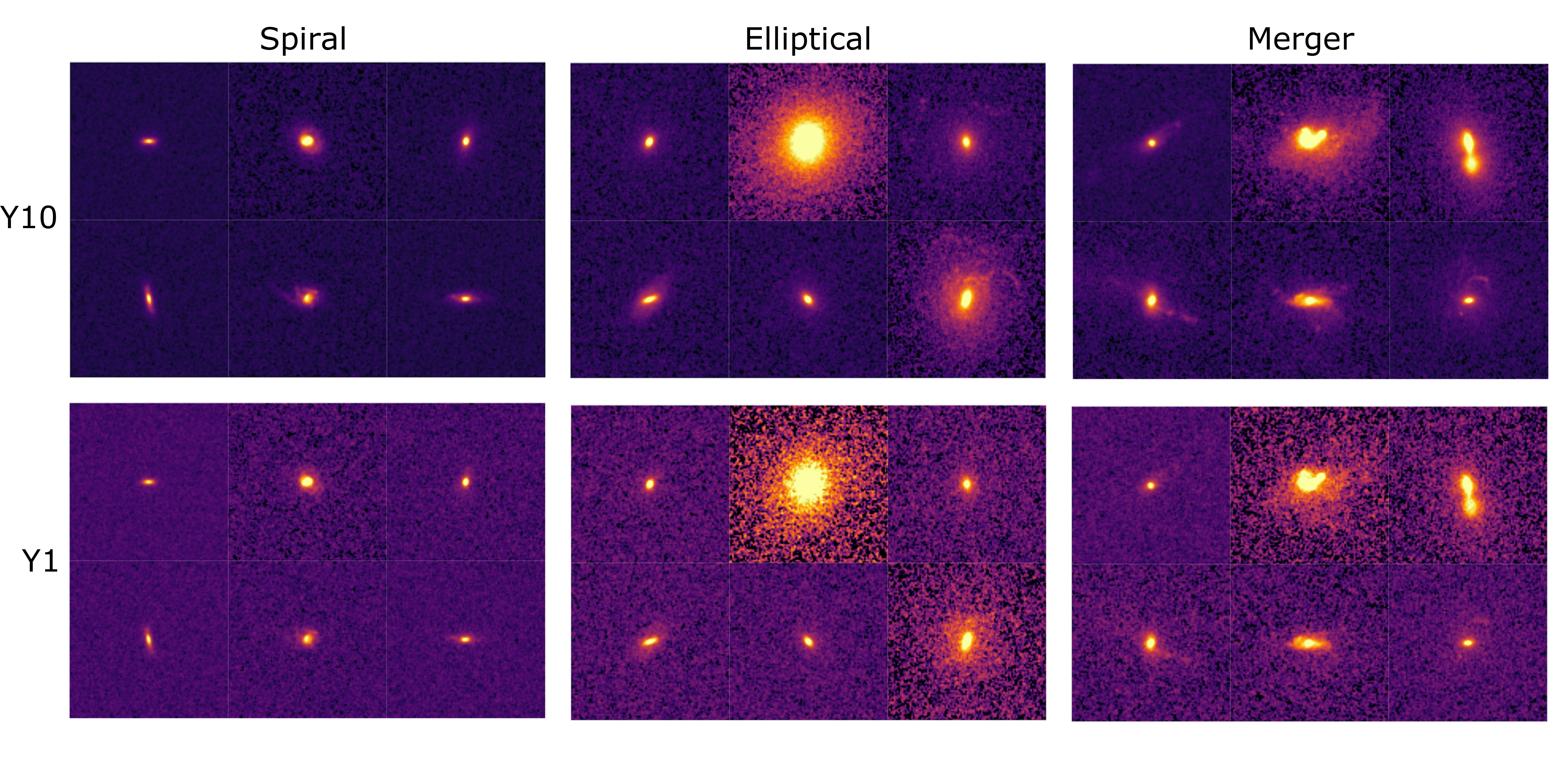}\\
    \caption{Example images from our datasets. 
    The left, middle, and right column show  examples of spiral, elliptical, and merging galaxies, respectively, with emulations of ten-year observational noise (Y10, top row), and one-year observational noise (Y1, bottom row).}
    \label{fig:data}
\end{figure*}

\subsection{Labeling classes}

To produce the labels for our experiments, we use the IllustrisTNG100 morphology catalogs~\citep{RS2019}, which include non-parametric morphological diagnostics, such as the relative distribution of the galaxy pixel flux values ~\citep[Gini coefficient $G$;][]{G1962}, the second-order moment of the brightest $20\%$ percent of the galaxy’s flux $M_{20}$~\citep{LP2004}, the concentration–asymmetry–smoothness ($CAS$) statistics~\citep{CB2003,LP2004}, and 2D S\'ersic fits~\citep{S1963}. 

We follow~\cite{LP2004} and~\cite{ST2015}, and use the $G$-$M_{20}$ ``bulge statistic'' to label spiral, elliptical, and merging galaxies.
Figure~\ref{fig:classes} presents the $G$-$M_{20}$ diagram of our dataset, with the intersecting lines representing the boundaries between the three classes.
Merging galaxies are those where $G > -0.14M_{20} + 0.33$, while non-mergers (including spirals and ellipticals) satisfy $G > 0.14M_{20} + 0.80$. 
Elliptical (spiral) galaxies have a Gini coefficient greater (lesser) than  $(0.693 M_{20} + 3.96)/4.95$.
The intersection of boundaries between the three classes lies at $(G_0, M_{20,0}) = (0.565, -1.679)$. 

From two IllustrisTNG100 snapshots (95 and 99), we extract $14,312$ spiral, $8,151$ elliptical, and $2,542$ merging galaxies.
To generate more data and to increase parity amongst the classes, we augment mergers with horizontal and vertical flips and with $90$-deg and $180$-deg rotations, producing 12,710 merger images.
We then divide these $\approx$35,000 images into training, validation, and test datasets with proportions $70:10:20$ (by randomly selecting images from the full dataset). 
For example images, see Figure~\ref{fig:data}.
  
\subsection{Perturbation: Noise}
\label{sec: noise}

To create our data, which emulates LSST observations, we use the \texttt{GalSim} package~\citep{RJ2015} and follow the same procedure as in~\cite{SM2021}.

We create two sets of survey-emulating images---a high-noise one-year survey (``Y1'') and a low-noise ten-year survey (``Y10'')---by applying an exposure time corresponding to one year or ten years of observations directly to the raw images ($552\,\mathrm{s}$ per year for $r$ and $i$ filters and $240\,\mathrm{s}$ for $g$ filter). 
This procedure simplifies data handling and obtains similar results to co-adding, where multiple single-epoch (30s) exposures are combined to yield the final results\footnote{Typical co-adding strategies consist of adding images using inverse variance weighting; in our case where the variance follows a perfectly known Poisson distribution, co-adding and simulating the full exposure are equivalent procedures.}. Furthermore, we incur PSF blurring for both atmospheric and optical PSF models. 
The images are simulated using a constant sky-background corresponding to the median sky level tabulated in~\cite{IK2019}. This background signal is subtracted from the final images containing the simulated Illustris sources, following the typical procedure used for real astronomical images. Thus, for the empty regions (without the simulated Illustris galaxy) on these images, we expect that the pixel levels follow a Poisson distribution centered at 0 and a variance equal to the original mean background level.

We then also process the images to make the details of galaxies more apparent, by clipping the pixel values to $0.1$ and $99.9$ percentiles, which removes a very small number of outlying pixels. 
We then perform arcsinh stretching to make fainter objects more apparent while preserving the original color ratios in each pixel, by scaling each of the three filters with $\mathrm{arcsinh}(c x)$, where $c=0.85$ is a constant used to scale the outputs to the range $[0, 1]$.

\subsection{Perturbation: One-pixel attacks}
\label{sec: one_pixel}

Multiple processes in astronomy data pipelines can change small number of pixels, including image (de)compression, errors in charge-coupled device (CCD) detectors, detector readout, and cosmic rays. We use the one-pixel attack as a proxy for these pixel-level perturbations.

To model one-pixel attacks, we represent the original image as a tensor $\bm{x}$ and its classification score as $p(\bm{x})$. 
An attack is optimized to find the additive perturbation vector $e(\bm{x})$ that maximizes the score $p_\mathrm{pert}(\bm{x}+e(\bm{x}))$ of the image for an incorrect class. 
The length of the perturbation vector must be less than a prescribed maximum: $\|e(\bm{x})\|_0 \leq L$, where $L=1$ for a one-pixel attack~\citep{SV2017}.

Creating an optimal attack is typically performed through differential evolution~\citep{SP1997,DS2011,SV2017}, a population-based optimization algorithm.
In each iteration of the algorithm, a set of candidate pixels (children) is generated according to the current population (parents) during each iteration.
To maintain population diversity, children are only compared to their corresponding parent and are kept if they possess a higher fitness value. 
For adversarial attacks, fitness is measured by the increase of the classification score for the desired incorrect class. 
The number of iterations required to find the optimal pixel-level perturbation corresponds to the susceptibility of a model to an attack.

\section{Networks and Experiments}
\label{sec: Networks}

We study the effects of perturbations in astronomical images in the context of two neural networks, that represent distinct levels of network complexity and sophistication. These networks are trained using labeled images to perform a supervised learning classification task of distinguishing between spiral, elliptical and merging galaxies.
Furthermore, we also explore the efficacy of domain adaptation for improving the performance and robustness of each of these networks.

\subsection{Network architectures}

\begin{table*}
  \centering
  \noindent\begin{minipage}[b]{\textwidth}
   \centering
    \caption{The architecture of the \textit{ConvNet} CNN used in this paper.}
  \label{table:arch}
  \centering
  \footnotesize
  \begin{tabular}{|l | l l l  l   l |}
\hline
\bf{Layers}         & \bf{Properties}           & \bf{Stride}       & \bf{Padding}  & \bf{Output Shape} & \bf{Parameters}   \\ \hline\hline
Input               & $3\times100\times100$\footnote{We use the ``channel first" image data format.}       &                  &              & (3, 100, 100)       & 0                 \\
Convolution (2D)    & Filters: 8                & $1$        & $2$          & (8, 100, 100)       & 608               \\
                    & Kernel: $ 5\times5$       &                  &              &                  &                 \\
                    & Activation: ReLU          &                  &              &                  &                  \\ \hline
Batch Normalization &                          &                 &              & (8, 100, 100)       & 16               \\ \hline
MaxPooling          & Kernel: $2\times2$        & $2$        & 0         & (8, 50, 50)       & 0                 \\ \hline
Convolution (2D)    & Filters: 16               & $1$        & $1$          & (16, 50, 50)      & 1168              \\
                    & Kernel: $ 3\times3$       &                  &              &                  &                  \\
                    & Activation: ReLU          &                  &              &                  &                  \\ \hline
Batch Normalization &                          &                  &              & (16, 50, 50)      & 32               \\ \hline
MaxPooling          & Kernel: $2\times2$        & $2$        & 0         & (16, 25, 25)      & 0                 \\ \hline
Convolution (2D)    & Filters: 32               & $1$        & 1          & (32, 25, 25)      & 4640              \\
                    & Kernel: $ 3\times3$       &                  &              &                  &                  \\
                    & Activation: ReLU          &                  &              &                  &                  \\ \hline
Batch Normalization &                          &                  &              & (32, 25, 25)      & 64                \\ \hline
MaxPooling          & Kernel: $2\times2$        & $2$        & 0         & (32, 12, 12)        & 0                 \\ \hline
Flatten             &                          &                  &              & (4608)            &                  \\ \hline
Bottleneck    &          &                  &              & (256)              & 1179904              \\ \hline
Fully connected     & Activation: Softmax       &                  &              & (3)               & 771                \\\hline
\multicolumn{2}{c}{} & \multicolumn{3}{l}{Total number of trainable parameters:} & \multicolumn{1}{l}{$1 186 432$}\\ 
\end{tabular}
\end{minipage}
\end{table*}

For a relatively simple model, we use a CNN that has three convolutional layers (with each layer followed by ReLU activation, batch normalization, and max pooling) and two dense layers; hereafter we refer to this model as \textit{ConvNet}. 
Details of the \textit{ConvNet} architecture are shown in Table~\ref{table:arch}.
For a more complex model, we use one of the smallest standard off-the-shelf residual neural networks, \textit{ResNet18}, which has four residual blocks (each containing convolutional layers), followed by two dense layers~\citep{HZ2016}.
Both networks have a latent space (layer immediately following the last convolution layer) of dimension 256, followed by an output layer with three neurons, one neuron corresponding to each of three classes: spiral, elliptical, and merging galaxies.
\textit{ConvNet} (\textit{ResNet18}) has ${\approx}$1.2M (${\approx}$11.2M) trainable parameters.
Training is performed by minimizing the weighted cross-entropy (CE) loss 

\begin{equation}
{\cal L}_\mathrm{CE}= \frac{- \sum\limits_{m=1}^{\mathrm{M}} w_m y_m \log \hat{y}_m}{\sum\limits_{m=1}^{\mathrm{M}} w_m},
\end{equation}
where the weight (distinct from the network weight parameters) for each class is calculated as $w_m = \frac{\mathrm{N}}{\mathrm{M} n_m}$, where $n_m$ is the number of images in class $m$, $\mathrm{M}=3$ is the total number of classes, and $N$ is the total number of images in the training dataset.

\subsection{Domain adaptation}
\label{sec: domain_adapt}
 
Domain Adaptation (DA) techniques help align the latent data distributions, allowing a model to learn the features shared between the two data domains and to perform well in both~\citep{C2017,WD2018,GD2020}. 
To align latent data distributions, we use Maximum Mean Discrepancy (MMD), which is a distance-based DA method that minimizes the non-parametric distance between mean embeddings of two probability distributions~\citep{SG2007,GB2008,CK2021}.
Generally, it is difficult to compare two probability distributions that are not completely known, but only sampled. To address this, in practice, kernel methods are used to map probability distributions into the higher-dimensional reproducing kernel Hilbert space.
This preserves the statistical features of the original probability distributions, while allowing one to compare and manipulate distributions using Hilbert space operations, such as the inner product. 

We follow~\cite{ZZ2020} and implement MMD as in~\cite{CK2021}, by using a combination of multiple Gaussian radial basis function kernels, $k(\theta,\theta')=\exp{-\frac{\|\theta-\theta'\|^2}{2\sigma^2}}$, where $\|\theta-\theta'\|$ is the Euclidean distance norm, $\theta$ and $\theta'$ are samples from any of the two latent data distributions, and $\sigma$ is the free parameter that determines the width of the kernel $k$, which measures similarity between two arguments $\theta$ and $\theta'$. In this work, we minimize the MMD distance between the Y10 and Y1 latent data distributions. 
We express the MMD loss as
\begin{equation}
\begin{split}
{\cal L}_\mathrm{MMD} = \frac{1}{\mathrm{N}(\mathrm{N}-1)} \sum
\limits_{i\neq j}^{\mathrm{N}}[& k(\theta_\mathrm{10}(i), \theta_\mathrm{10}(j))\,- \\
&k(\theta_\mathrm{10}(i), \theta_\mathrm{1}(j))\,- \\
&k(\theta_\mathrm{1}(i), \theta_\mathrm{10}(j))\,+ \\
&k(\theta_\mathrm{1}(i), \theta_\mathrm{1}(j))],
\end{split}
\end{equation}
where $\mathrm{N}$ is the total number of training samples $\theta_\mathrm{10}$ from the Y10 or $\theta_\mathrm{1}$ from the Y1 latent data distribution (in our dataset, both distributions have the same number of samples $\mathrm{N}$). 
For more details about the MMD distance calculation, see~\citep{SG2007,GB2008,CK2021}.

When using DA, the total loss ${\cal L}_\mathrm{TOT}$ is composed of the MMD and the CE loss:
\begin{equation}
{\cal L}_\mathrm{TOT} = {\cal L}_\mathrm{CE} + \lambda {\cal L}_\mathrm{MMD},
\end{equation}
where $\lambda \geq 0$ controls the relative contribution of the MMD loss.
The MMD performs domain adaptation and alignment of the two latent data distributions by calculating kernel values for all possible combinations of the latent space embeddings (for objects from the same dataset, as well as cross-dataset). The minimization of the MMD loss requires the maximization of the kernels $k(\theta_\mathrm{10}, \theta_\mathrm{1})$ and $k(\theta_\mathrm{1}, \theta_\mathrm{10})$ that describe cross-similarities between the two data distributions. 
This results in the model being forced to find domain-invariant features, which make cross-similarities large.

During training, the CE loss requires labeled images. In our experiments (both regular training without domain adaptation and with domain adaptation), networks are trained using our baseline low-noise Y10 images and corresponding labels (spiral, elliptical or merger). On the other hand, the MMD loss uses only latent space image embeddings from both Y10 and Y1 datasets and does not require labels. This feature of the MMD loss is particularly valuable in cases when one of the datasets is unlabeled.

\subsection{Hyperparameters and training}

We use the Adam optimizer~\citep{KB2014} with beta values of $(\beta_1,\beta_2)=(0.7, 0.8)$ and a weight decay (L2 penalty) of $0.001$ for regular training and $0.0001$ for domain adaptation training.
The initial learning rate in all our experiments is $0.00001$. 
We use fixed batch sizes of 128 during training and 64 during validation and testing.
The training length is set to 100 epochs, but we use early stopping to prevent overfitting. 
When using domain adaptation, through experimentation with various values, we set $\lambda = 0.05$ for the MMD loss term.
When shuffling images and initializing network weights for training, we ensured consistency of results by setting one fixed random seed ($0$) for all experiments.
Training was performed on an Nvidia Tesla V100 GPU in Google Cloud. 

\section{Assessing Model Robustness}
\label{sec: latent}

We assess the robustness of trained neural networks when they are presented with data that has been perturbed. 
First, we employ the simple network classification accuracy and other standard performance metrics.
Next, we study the distributions of the distances between original and perturbed data in the latent space.
Then, we visualize the trained latent space using two techniques: \textit{church window plots} that show specific directions in the latent space; and \textit{isomaps} that show lower-dimensional projections of the latent space.

\subsection{Distance metrics}
\label{sec: dist}

Perturbations to images move their positions within the network's trained latent space, which can cause an object to cross a decision boundary from the region corresponding to the correct class to a region corresponding to the incorrect class.
If a method increases the model robustness to perturbations, crossing the decision boundary and entering the wrong class region will require the image to move further from its origin. In other words, the region of the wrong class will become further away from correctly classified images. 

We randomly select a 150-image sub-sample of our test dataset on which to apply one-pixel perturbations.
This sample is large enough for statistically significant characterization of distances between the perturbed and unperturbed data distributions, and it is small enough to generate a one-pixel attack and run visual inspection on all the images.
We then choose the images that were successfully flipped for both regular and domain adaptation training, which amounts to 136 for \textit{ResNet18}.

We use two distance metrics to quantify the sensitivity of our models to perturbations and compare  latent spaces of models trained without (regular training) and with domain adaptation. First, for each image in the baseline dataset and its perturbed counterparts, we calculate the Euclidean distance $d_\mathrm{E}$ between the latent space positions of the baseline and the perturbed images. Next, we calculate the Jensen-Shannon (JS) distance~\citep{L1991} between the distributions of Euclidean distances $d_\mathrm{E}$, for models trained using regular training and training with domain adaptation. 
The JS distance is the square root of the JS divergence, which is a measure of similarity between two probability distributions~\citep{L1991}. The JS divergence is a symmetrized and smoothed version of the well-known Kullback–Leibler divergence $D_\mathrm{KL}(P\parallel Q)$~\citep{KL1951} and can be calculated as:
\begin{equation}
    \mathrm{JS}(P\parallel Q) = \frac{1}{2} D_\mathrm{KL}(P\parallel R) + \frac{1}{2} D_\mathrm{KL}(Q\parallel R),
\end{equation}
where $R=\frac{1}{2}(P+Q)$ and $P$ and $Q$ are the two probability distributions.

\subsection{Perturbation direction: Church window plots}
\label{sec: church}

We also seek to investigate how a perturbation in an image affects its latent space representation and thus classification.
Church window plots, named after the often-colorful stained glass windows, visualize the latent space regions for classes in the proximity of a given image~\citep{GS2014,H2017,FG2019}.

First, in a plot, we place the latent space embedding of the unperturbed baseline image at the origin. 
Then, we subtract the unperturbed image's latent embedding from that of the perturbed image, yielding the latent space representation of the perturbation vector. We chose to orient the plane such that the horizontal axis lies along the one-pixel perturbation direction, and the vertical axis lies along the noisy direction; in principle any perturbation direction can be chosen.
In our plots, we take a slice of the entire latent space, motivated by the desire to visualize the model behavior in the direction of perturbations we chose for basis vectors. 

Next, the perturbation vectors are discretized into small steps in each direction. 
All possible combinations of these perturbations are added to the baseline image to create new perturbed image embeddings. 
These new embeddings are then passed into a truncated network consisting of only the dense layers of our original trained model: a $256$-dimensional layer and an output layer with three neurons.
This truncated network necessarily shares the same weights as the flattened layers of the original network, and outputs the classification result of the given perturbed image embedding. 
This classification determines the color of that pixel on the plot.

A church window plot shows relative distance; each axis is normalized to $[-1,1]$ based on the latent space representation for that image.
Therefore, it is difficult to use such plots to compare church window representations for different images.
We deviate slightly from traditional church window plot applications, e.g.~\citep{H2017}, wherein the authors oriented the horizontal axis with the adversarial (perturbation) direction, while the other axis is calculated to be orthonormal; we instead have two perturbation directions.
Also, traditionally, the color white is used to designate the correct class.

\subsection{Low-dimensional projections with isomaps}
\label{sec: isomaps}

Next, we project our high-dimensional latent spaces to two and three dimensions, which is nontrivial. 
Linear projections, such as those generated by Principal Component Analysis ~\citep[PCA;][]{P1901}, often miss important non-linear structures in the data. 
Alternatively, manifold learning respects non-linear data patterns; some example algorithms are t-distributed stochastic neighbor embedding ~\citep[tSNE;][]{MH2008}, locally linear embedding~\citep{RS2000}, and the isomap~\citep{TS2000}. 

In this work, we use the isomap, which is a lower-dimensional embedding of a network latent space, such that geodesic distances in the original higher-dimensional space are also respected in the lower-dimensional space.
The isomap-generation algorithm has three major stages. 
First, a weighted neighborhood graph $G$ over all data points is constructed,  either by connecting all neighboring points that are within some chosen radius $\epsilon$ or by selecting data points among the $K$ nearest neighbors. 
We used the \texttt{scikit} implementation of isomaps, which has an option for \texttt{auto} that instructs the algorithm to select the optimal method for graph construction~\citep{PV2011}.
Within graph $G$, the edge weight values are assigned the distances between neighboring points. 
Next, the geodesic distances between all pairs of points on the manifold are estimated as their shortest-path distances in the graph $G$. 
Finally, the lower $d$-dimensional embedding that best preserves the manifold’s estimated intrinsic geometry (low-dimensional representation of the data in which the distances respect well the distances in the original high-dimensional space) is produced by applying classical Metric Multidimensional Scaling ~\citep[MDS;][]{BG2005} to the matrix of the shortest-graph distances. Most commonly, $d$ is $2$ or $3$, to facilitate visualization.

\section{Results}
\label{sec: results}

We assess the performance and robustness of the two deep learning models, each trained without (regular training) and with domain adaptation.
We train on data in one domain (Y10) and test on data that has been perturbed in one of two ways: with a one-pixel attack (1P) representing data processing errors, or with inclusion of higher observational noise (Y1).

\subsection{Classification accuracy and other network performance metrics}

We focus first on the more complex \textit{ResNet18} network, which achieved higher classification accuracy. The setup, results, and slight differences in behavior for the simpler \textit{ConvNet} model are discussed in Section~\ref{sec: ConvNet}. Overall, we find that both models respond similarly to image perturbations and exhibit improved performance when domain adaptation is used during training.

Without domain adaptation, the \textit{ResNet18} model achieves an accuracy of $72\%$ ($43\%$) when tested on Y10 (Y1) images.
When we use domain adaptation, the accuracy is $82\%$ ($66\%$) when tested on Y10 (Y1) images.
Using domain adaptation prevented the more complex model from overfitting, which helps increase the accuracy in the final epoch on the baseline Y10 images; this was also observed in~\cite{CK2021}. Domain adaptation also helped increase the accuracy on noisy Y1 data by $23\%$.
In Table~\ref{table:performance}, we report the accuracy, as well as weighted precision, recall, and F1 score for \textit{ResNet18} regular training and training with domain adaptation.
Weighted metrics are calculated for each of the three class labels, and then their average is found and weighted by the number of true instances for each label.

\begin{table}
   \centering
   \noindent\begin{minipage}[b]{0.99\columnwidth}
   \centering
    \caption{
    Performance metrics for \textit{ResNet18} on Y10 and Y1 test data for regular training (top row) and training with domain adaptation (bottom row). 
    The table shows the accuracy and weighted precision, recall, and F1 scores. 
    Domain adaptation increases performance in all metrics for both Y10 and Y1 data.
    }
  \label{table:performance}
  \centering
  \begin{tabular}{|l || l |c c|}
 \hline Training      &   Metric   &  Y10  & Y1  \\\hline \hline
\multirow{4}{*}{Reg}               &  Accuracy     &   $0.72$       &  $0.43$    \\ 
                                                &  Precision    &   $0.76$       &  $0.61$   \\
                                                &  Recall       &   $0.72$       &  $0.43$   \\
                                                &  F1 Score     &    $0.72$      &  $0.36$   \\\hline
\multirow{4}{*}{DA}         &  Accuracy     &   $0.82$       &  $0.66$   \\ 
                                                &  Precision    &   $0.82$       &   $0.67$   \\
                                                &  Recall       &   $0.82$       &   $0.66$     \\
                                                &  F1 Score     &   $0.82$       &   $0.67$  \\\hline
\end{tabular}
\end{minipage}
\end{table}

\subsection{Case studies: Latent space visualizations of perturbed data }
\label{sec: effects}

\begin{figure*}[!ht]
   \centering
	\includegraphics[width=\linewidth]{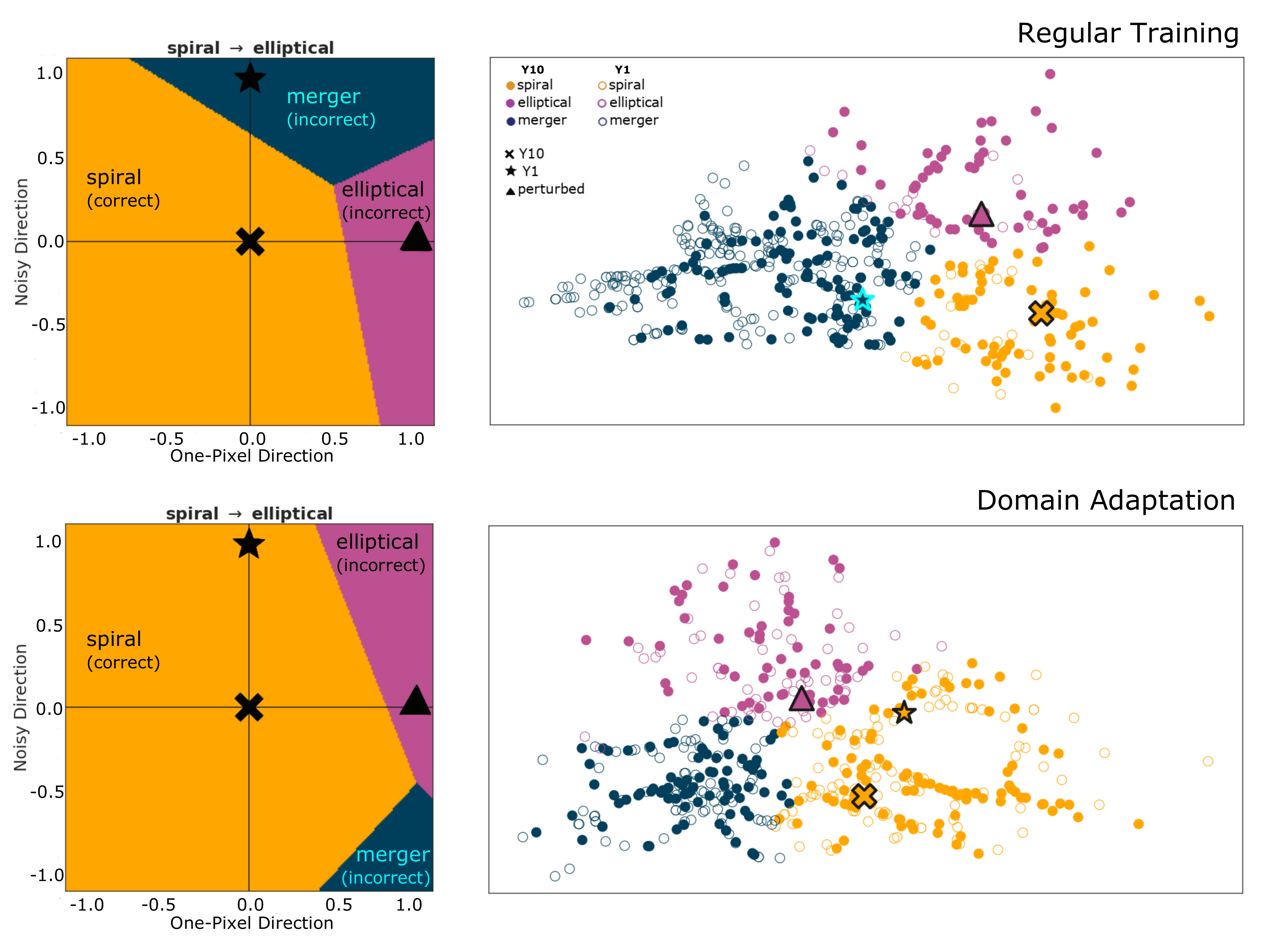}\\
    \caption{
    Church window plots (left) and isomaps (right) of an example triplet of a baseline Y10 image (``$\protect\mathbf{\times}$'') with each of the two perturbations: noisy Y1 (star) and one-pixel attack (triangle)~\protect\footnotemark.
    The top row of images corresponds to the model trained without domain adaptation (regular training), while the bottom row shows the same triplet of images for a model trained with domain adaptation.
    In all plots, classification into spiral galaxies is shown in orange, elliptical in violet, and merger in navy blue. 
    Isomaps are constructed from 250 randomly selected images from our test set, with Y10 images shown as filled circles, and Y1 as empty circles. 
    Each church window plot is labeled with the true class of the Y10 image and the class targeted by the one-pixel attack: ``true class $\protect\rightarrow$ targeted class''. 
    Most noisy Y1 images are incorrectly classified as mergers when training without domain adaptation, and thus most Y1 points in the top isomap are navy blue. 
    Adding domain adaptation improves the overlap between the Y10 and Y1 data distributions, which leads to correct classification of all three classes in both the Y10 and Y1 datasets. The one-pixel attacked image is incorrectly classified by both the regular training and domain adaptation models, but more differential evolution iterations were needed to find such a pixel for the domain adaptation model.}
    \label{fig:results}
\end{figure*}

Next, we investigate the classification of a single spiral galaxy image in three forms---the baseline and the two perturbations---by visualizing the network latent space representation of each form.
Figure~\ref{fig:results} presents church window plots and 2D isomaps of latent space representations given a \textit{ResNet18} network with regular training (top) and with DA training (bottom).
The church window panel shows that with regular training, the one-pixel attack moved the latent space representation into the elliptical region, while the noise moved the representation to the merger region.

The isomap panel shows a 2D projection of the latent space representation for 250 randomly selected objects in our test dataset, as well as the three forms of our single example galaxy: Y10 (``$\times$''), Y1 (star), 1P (triangle).
The filled (empty) circles show the Y10 (Y1) latent representation of the randomly selected galaxies (we pick the same galaxies from both Y10 and Y1 data).
For the baseline Y10 dataset, which the model was trained on, examples are clearly separated into three classes---spiral (orange), elliptical (violet), and merger (navy blue)---for both regular and domain adaptation training. 

With the regular training, $88\%$ of the Y1 data shown on the isomap are incorrectly classified as mergers (empty navy blue circles).  
Using domain adaptation training produces a clear class separation in the Y1 data as well (with the accuracy on the Y1 test set increasing by $23\%$), leading to good overlap between the Y1 and Y10 classes and a common decision boundary, as we can see on the isomap in the bottom row. 
Both the church window plot and the corresponding isomap show that with domain adaptation the example Y1 image from the triplet is correctly classified as a spiral galaxy. 
The one-pixel attack still manages to flip the Y10 image to elliptical, but because the incorrect class region is now further away, more iterations of differential evolution were needed (see Section~\ref{sec: distances} for details). 

\begin{figure*}
	\includegraphics[width=\linewidth]{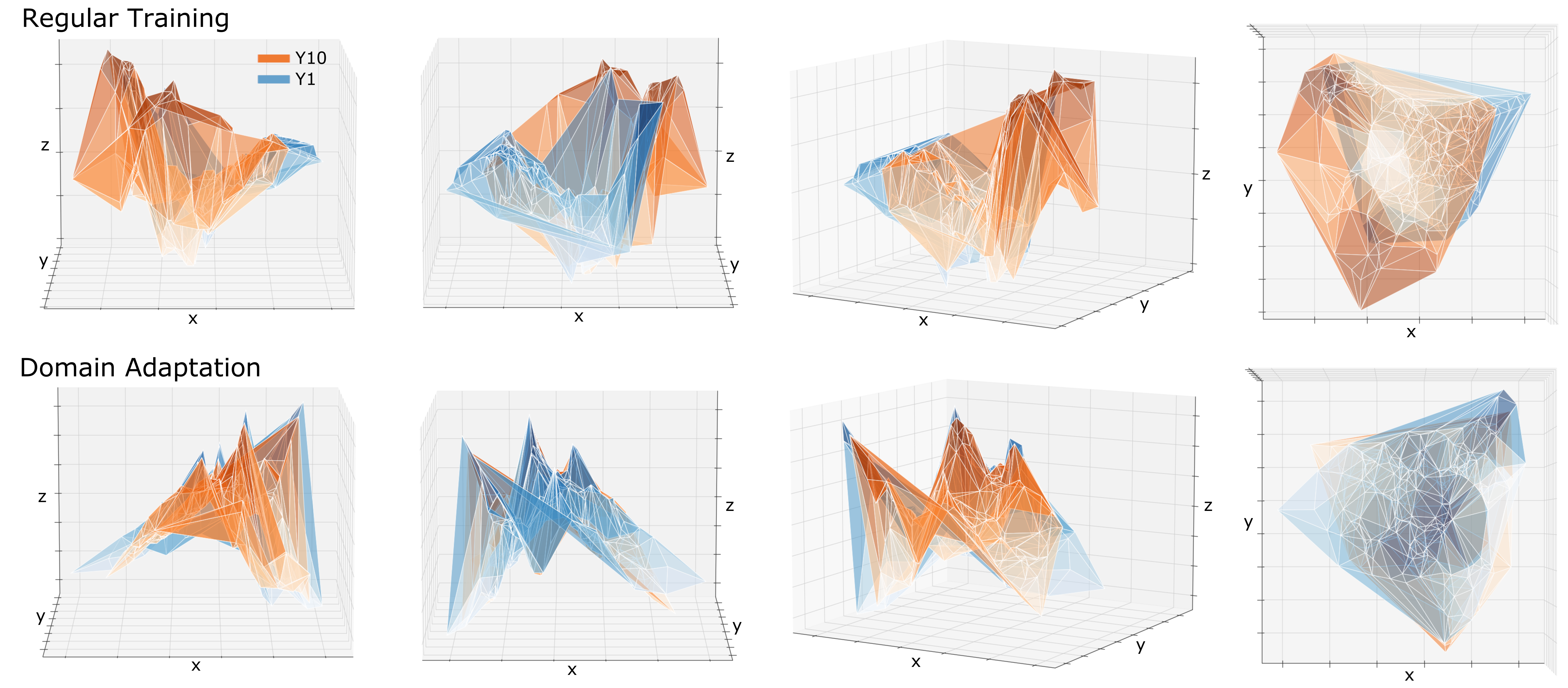}
    \caption{
    3D isomap projections of the 256-dimensional latent space, derived from 200 randomly sampled data points from test sets,
    in four camera orientations (regular training in top row, domain adaptation in bottom row). 
    The baseline Y10 images are connected with an orange plane, while the noisy Y1 images are connected with a blue plane. 
    Domain adaptation helps increase the overlap of the Y10 and Y1 distributions.}
    \label{fig:isomaps3d}
\end{figure*}

To understand how the data moves in the latent space after domain adaptation is employed, Figure~\ref{fig:isomaps3d} shows illustrative 3D isomaps of Y10 and Y1 test data\footnote{To further illustrate the differences between Y10 and Y1 latent data distributions, we show videos of rotating 3D isomaps (made from 50 randomly chosen test set images, due to memory constraints) as supplementary online material, as well as on our \href{https://github.com/AleksCipri/DeepAdversaries}{GitHub page}.}.
Here, we can clearly see that without domain adaptation (top row), the noisy Y1 data is not overlapping with the Y10 data. 
In fact, Y1 data is concentrated in a small region of the plot. 
On the other hand, with the inclusion of domain adaptation (bottom row), both the Y10 and Y1 data distributions follow the same trend and data distributions overlapping quite well. 

\begin{figure*}
   \centering
	\includegraphics[width=\linewidth]{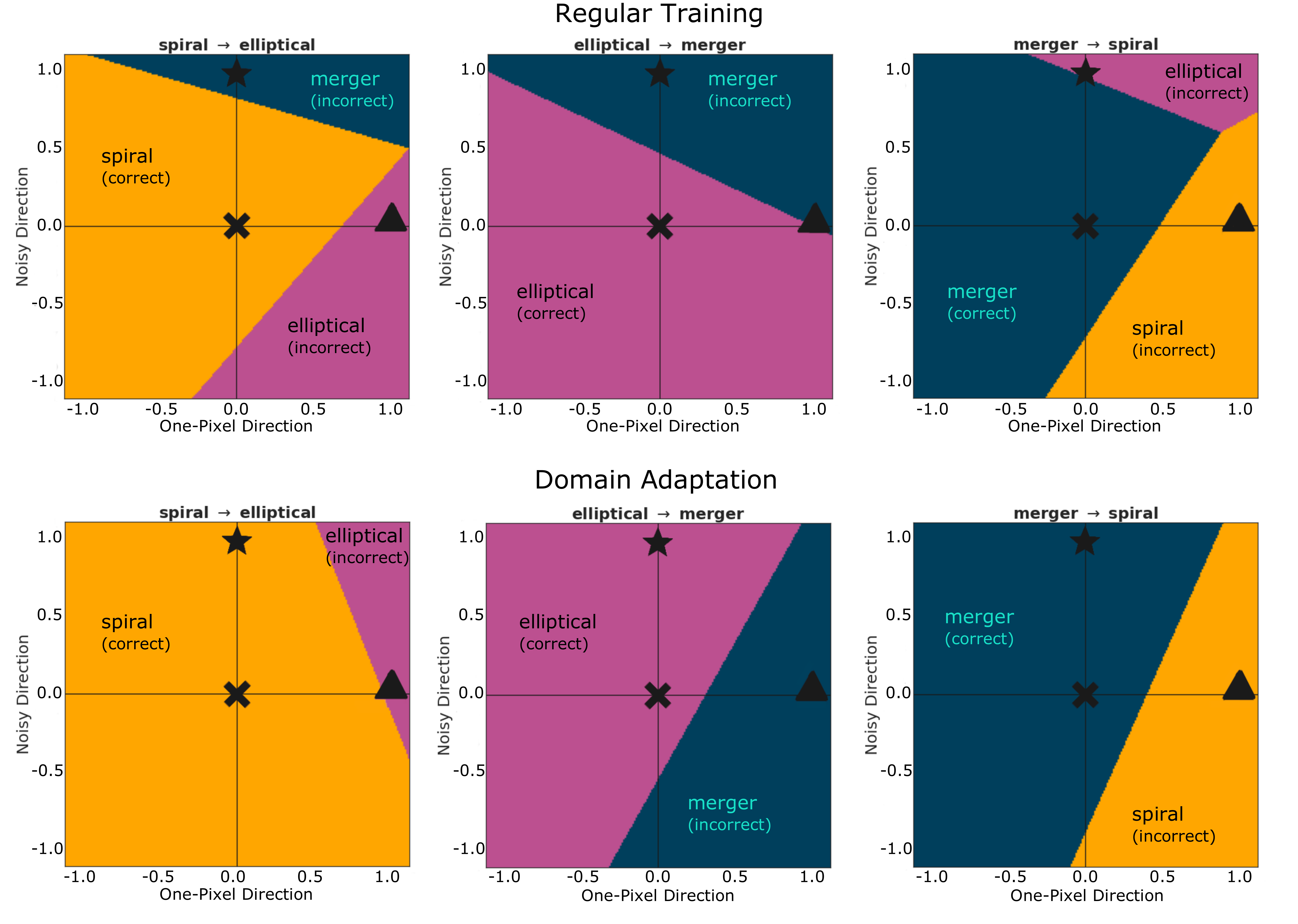}\\
    \caption{Example church window plots for \textit{ResNet18}. Each plot shows an example triplet of a baseline Y10 image (``$\protect\mathbf{\times}$'')  with each of the two perturbations: noisy Y1 (star) and one-pixel attack (triangle).
    The top (bottom) row shows the same three examples classified by the model trained using regular (domain adaptation) training.
    We selected examples for which the Y10 image was correctly classified after training, and for which the one-pixel attack is successful.
    Each church window plot title includes the true class of the baseline Y10 image and the targeted incorrect class for the pixel-attacked image: ``true class $\rightarrow$ targeted class.'' We emphasize that church window plot shows relative distances, based on the latent space representation for a particular image with each axis normalized to $[-1,1]$, so distances should not be compared between different plots. While the model trained with regular training incorrectly classifies both noisy Y1 and one-pixel attacked images, the model trained with domain adaptation is correct for both Y10 and Y1 data. The one-pixel attack was still successful with domain adaptation, but more differential evolution iterations were needed to find such a pixel.
    }
    \label{fig:church_RN}
\end{figure*}

Figure~\ref{fig:church_RN} shows additional \textit{ResNet18} church window plots for several examples from the 150-image test sub-sample, for which we performed the one-pixel attack. 
The top and bottom rows show the same triplet examples for regular and domain adaptation training, respectively.
The first row shows examples of baseline Y10 images that were correctly classified (spiral, elliptical, merger) and then successfully flipped to a different class with a one-pixel attack. 
As shown in the top row of images, when regular training is used, the one-pixel perturbed, the noisy Y1, and the baseline Y10 image can each belong to three different classes. On the other hand, when domain adaptation is used (bottom row), both Y10 and Y1 examples are correctly classified; the church window plots often only show two of the three possible class regions.
Domain adaptation leads to more robustness and higher output probabilities, which also means that the one-pixel attack needs to move the image further in the latent space in order to reach the region of the wrong class, so more iterations of differential evolution are needed to find such a pixel. 
We limit the differential evolution procedure, which seeks the adversarial pixel, to 80 iterations. Within the maximum number of iterations only 136 images (out of the 150-image test set sub-sample) were successfully flipped to the wrong class after the inclusion of domain adaptation. We emphasize that the church window plot shows relative distance: each axis is normalized to $[-1,1]$ based on the latent space representation for that image. Hence, distances should not be compared between different church window plots. Section~\ref{sec: distances} includes quantitative comparisons of distance metrics.

\begin{figure*}
   \centering
	\includegraphics[width=\linewidth]{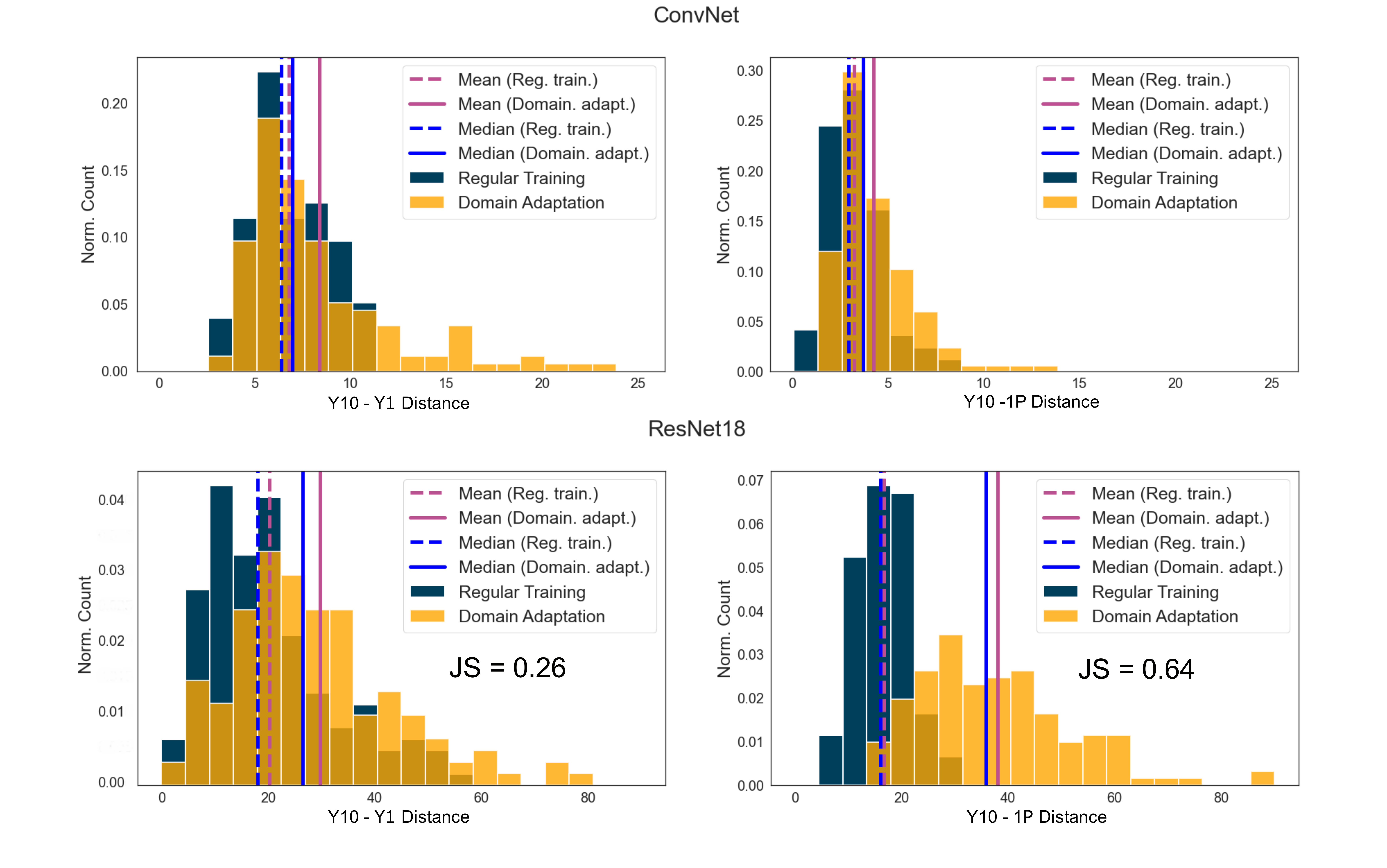}\\
    \caption{Distributions of Euclidean distances between Y10 and noisier Y1 images (left), and between Y10 and one-pixel (1P) perturbed images (right), for the \textit{ConvNet} (top) and \textit{ResNet18} (bottom) models.
    Distances (given in Table~\ref{table:distancesRN} and Table~\ref{table:distancesCN}) for the regular training are plotted in navy blue, with mean (violet) and median (blue) values plotted as dashed lines. 
    Distances for the model trained with domain adaptation are plotted in orange, with mean (violet) and median (blue) plotted as solid lines. 
    On each plot we also report the JS distance as a measure of the difference between the two distributions. 
    In our experiments, domain adaptation increases the distances to all types of perturbations. In particular, the increase in the JS distance between the regular and the successfully flipped one-pixel perturbed images (right column) is a good indicator of the increased model robustness.
}
    \label{fig:hist_RN}
\end{figure*}

\subsection{Distances metrics: Characterizing network robustness with latent space distances}
\label{sec: distances}

We use Euclidean distances $d_\mathrm{E}$ between baseline Y10 images and their perturbed counterparts in the network latent space to assess model robustness. First, we estimate the median, the mean and standard errors of the distribution of distances between baseline Y10 images and their perturbed counterparts in the latent space of the \textit{ResNet18} model in Table~\ref{table:distancesRN}.
For both types of perturbations (noisy and one-pixel attack), we observe that domain adaptation increases the distance $d_\mathrm{E}$ between the baseline and perturbed images. 

\begin{table}
\caption{
Medians, means and standard errors of Euclidean distances in the latent space of \textit{ResNet18} for the $136$-image sub-sample of our test set of images. 
Domain adaptation increases median and mean distance to the one-pixel perturbed image, making the model more robust against this kind of attack (we also plot histograms of all distances in the bottom row of Figure~\ref{fig:hist_RN}).
}
\label{table:distancesRN}
\centering
\begin{tabular}{|l | l || c c c|}
 \hline \multirow{2}{*}{Perturb.}      &    \multirow{2}{*}{Training}     &      \multicolumn{3}{c|}{$d_\mathrm{E}$}    \\
 &  &  Median & Mean & St. Err.\\\hline \hline
\multirow{2}{*}{Y10 -- Y1}                    &  Reg            &    $18.1$     & $20.2$   &  $\pm 1.0$  \\ 
                                              &  DA                 &      $26.4$  & $29.7$ & $\pm 1.5$  \\\hline
\multirow{2}{*}{Y10 -- 1P}             &  Reg            &     $16.0$   & $16.8$   & $\pm 0.4$   \\ 
                                              &  DA                 &    $35.8$   & $38.1$ & $\pm 1.3$  \\\hline 
\end{tabular}
\end{table}

For Y10-Y1 distances, this happens because domain adaptation allows the Y1 data to align with all three classes and to be correctly classified, instead of being concentrated in one region with no class distinction. In top row of Figure~\ref{fig:results}, the isomap shows that regular training places all Y1 data very close to the Y10 merger class region. When domain adaptation is used all three classes from both datasets are correctly aligned, which ultimately leads to the increase in the mean Euclidean distance between Y10 and Y1 data. However, caution should be applied, because an increase in this distance will not necessarily happen for all datasets and neural network models. The change in this distance depends strongly on the location of the unknown dataset in the latent space of the model trained with regular training (which can be very unpredictable). If the unknown dataset is placed far away from the known data, including domain adaptation will reduce the distance between latent elements from the two domains. 

\footnotetext{On the top isomap, the star symbol is plotted with a cyan border to make it more visible, since it is located in the region with many other navy blue points.}

More importantly, when domain adaptation is included, the mean distance between Y10 and the one-pixel perturbed images that were incorrectly classified also increases (see Figure~\ref{fig:hist_RN} and Table~\ref{table:distancesRN}). This means that with domain adaptation, images that were successfully flipped to the wrong class needed to move farther to cross the class boundary and end up in the wrong class.
For the better performing \textit{ResNet18} model, mean $d_\mathrm{E}$ increases by a factor of ${\approx}2.3$, and median by ${\approx}2.2$, which means that the inadvertent data perturbations are less likely to successfully move the image to the wrong class. In this case, $d_\mathrm{E}$ is  the distance between correctly classified images and incorrect class regions: therefore, when $d_\mathrm{E}$ increases, so does the model robustness.

We also study the distributions of Euclidean distances between the baseline and the perturbed images under our two scenarios -- regular training and domain adaptation training. 
We normalize these distributions to sum to 1.
We then calculate the JS distance between the $d_\mathrm{E}$ distributions obtained with regular training and those obtained through training with domain adaptation. 
We illustrate the \textit{ResNet18} distributions in the bottom row of Figure~\ref{fig:hist_RN}. 
As with the $d_\mathrm{E}$ distribution means, the success of domain adaptation in increasing the distance to the one-pixel perturbed images for \textit{ResNet18} results in the larger JS distance of $0.64$. 

We emphasize here that this study was done using only $136$ images ($134$ for \textit{ConvNet}) that were successfully flipped by the one-pixel attack (due to computational constraints). For more precise quantification of the model behavior, a larger sample is needed. Still, even this small sub-sample shows trends in the model behavior and the potential benefit of using domain adaptation.

\subsection{ConvNet Results}
\label{sec: ConvNet}

Our simpler \textit{ConvNet} model architecture is presented in Table~\ref{table:arch}.

\textit{ConvNet} reaches slightly lower accuracies compared to \textit{ResNet18}, but exhibits similar behavior when trained with and without domain adaptation. 
With the regular training, the one-pixel attack more easily flips the image to an incorrect class, and most of the noisy Y1 images are incorrectly classified. 
When domain adaptation is employed, classification accuracy on noisy images increases by $9\%$, and successful one-pixel attacks are harder to find (more iterations of differential evolution are needed and the successfully attacked images are further away from the baseline Y10 image). 

Table~\ref{table:performanceCN} provides detailed metrics for the performance of \textit{ConvNet} on the Y10 and Y1 test data. One notable difference, compared to the more complex \textit{ResNet18}, is the slight drop in performance in the source domain when domain adaptation training is used (accuracy is lower by $1\%$). With domain adaptation, the model is forced to use only domain-invariant features, which makes the classification slightly harder in the source domain. Still, this is acceptable because these domain-invariant features allow the model to classify the target domain, which was not possible with regular training. On the other hand, more complex models like \textit{ResNet18} need to be trained more carefully, often with early stopping in order to prevent overfitting on the training dataset during the regular training. When domain adaptation is employed, it acts as a regularizer and allows the model to train for longer and improve its performance even in the source domain.

Furthermore, in Table~\ref{table:distancesCN} we give means and standard errors of Euclidean distances $d_\mathrm{E}$ between baseline Y10 images and noisy Y1 or one-pixel perturbed (1P) images, calculated for the 134-image sub-sample of the test set of images (images that were successfully flipped for both regular and domain adaptation training). In the top row of Figure~\ref{fig:hist_RN}, we plot distributions of these Euclidean distances and give the JS distance as a measure of the difference between the regular and domain adaptation distributions. Similar to \textit{ResNet18}, the simpler \textit{ConvNet} also exhibits improved robustness when trained with domain adaptation, which is reflected in $\approx1.3$ times larger mean and median Euclidean distance between baseline Y10 images and their perturbed counterparts and the increased classification accuracy on noisy Y1 data.

To compare distances in spaces with different dimensions (objects becoming more distant as the dimensionality grows), we require that both the \textit{ResNet18} and simpler \textit{ConvNet} have the same 256-dimensional latent space. 
Therefore, any different behavior of data points in this space can be attributed to different features the two networks find as important and exploit to build their latent spaces. It is important to keep in mind that these differences are a consequence of the vastly different model sizes (number of tunable parameters), as well as the type of the model (one a regular CNN and the other containing residual blocks).
Because the latent spaces of the two networks are different, we cannot directly compare the distance metrics for the \textit{ConvNet} and \textit{ResNet18} models. 
For this reason, we look at the overall behavior and the changes introduced by domain adaptation, in combination with church window plots and isomaps, to get a better understanding of the effects of image perturbations on the model performance. These results show the power of domain adaptation as a tool for increasing model robustness.

\begin{table}
   \centering
   \noindent\begin{minipage}{0.99\columnwidth}
   \centering
    \caption{
     Performance metrics for \textit{ConvNet} on Y10 and Y1 test data for regular training (top row) and training with domain adaptation (bottom row). 
    The table shows the accuracy and weighted precision, recall, and F1 scores.}
  \label{table:performanceCN}
  \centering
  \begin{tabular}{|l || l |c c|}
 \hline Training       &   Metric   &  Y10  & Y1  \\\hline \hline
\multirow{5}{*}{Reg}               &  Accuracy     &   $0.70$       &  $0.48$    \\ 
                                                &  Precision    &   $0.72$       &  $0.67$   \\
                                                &  Recall       &   $0.70$       &  $0.48$   \\
                                                &  F1 score     &    $0.70$      &  $0.42$  \\\hline
\multirow{5}{*}{DA}              &  Accuracy     &   $0.69$       &  $0.57$   \\ 
                                                &  Precision    &   $0.70$       &   $0.62$   \\
                                                &  Recall       &   $0.69$       &   $0.57$     \\
                                                &  F1 score     &   $0.69$       &   $0.58$  \\\hline
\end{tabular}
\end{minipage}
\end{table}

\begin{table}
\caption{
Medians, means and standard errors of Euclidean distances in the \textit{ConvNet} latent space. 
Values are calculated for the $134$-image sub-sample of our test set of images. 
Domain adaptation increases median and mean distance to the one-pixel perturbed image, making the model more robust against this kind of attacks (the top row of Figure~\ref{fig:hist_RN} shows histograms of all distances).
}
\label{table:distancesCN}
\centering
\begin{tabular}{|l | l || c c c|}
 \hline \multirow{2}{*}{Perturb.}      &    \multirow{2}{*}{Training}     &      \multicolumn{3}{c|}{$d_\mathrm{E}$}    \\
 &  &  Median & Mean & St. Err.\\\hline \hline
\multirow{2}{*}{Y10 -- Y1}                    & Reg            &       $6.3$      & $6.7$ & $\pm0.2$   \\ 
                                              &  DA                 &    $6.9$    & $8.3$ & $\pm0.3$    \\\hline
\multirow{2}{*}{Y10 -- 1P}             &  Reg            &   $2.8$    & $3.1$ & $\pm0.1$    \\ 
                                              &  DA                 &   $3.6$ & $4.1$ & $\pm0.2$     \\\hline 
\end{tabular}
\end{table}

\section{Discussion and Conclusion}
\label{sec: DandC}

In this paper, we explored how data perturbations that arise from astronomical processing and analysis pipelines can degrade the capacity of deep learning models to classify objects.
We then explored the efficacy of particular visualization techniques (church window plots and isomaps) in assessing model behavior and robustness in these classification tasks.  
Finally, we tested the use of domain adaptation for mitigating model performance degradation.

Our work focuses on the effects of two types of perturbations:
observational noise and image processing error (represented by the one-pixel attack).
We demonstrated that the performance of standard deep learning models can be significantly degraded by changing a single pixel in the image. 
Additionally, images with different noise levels (even if the noise model is the same) are also incorrectly classified if the model is only trained on one of the noise realizations. Even larger discrepancies between data distributions can arise if the two datasets include noise that cannot be described with the same model.

We illustrated how training on multiple datasets with the inclusion of domain adaptation leads to extraction of more robust features that can substantially improve performance on both datasets (Y1 and Y10). 
In other words, older high-noise (Y1) data can be used in combination with domain adaptation during training to increase performance and robustness of models intended to work with newer low-noise data (Y10). Furthermore, the added benefit of this type of training is increased robustness to inadvertent one-pixel (1P) perturbations that can arise in astronomical data pipelines.
We showed that the inclusion of domain adaptation during the training of \textit{ResNet18} increases the classification accuracy for Y10 data by $10\%$ (domain adaptation acts as a regularizer, allowing the model to train for longer and improve even in the source domain), while the accuracy in the noisy Y1 domain (which could not be classified at all without domain adaptation) increases by $23\%$.
Furthermore, to successfully flip an image to the wrong class using the one-pixel attack, the image needs to move ${\approx}2.3$ times further in the neural network's latent space after the inclusion of domain adaptation. In case of the simpler \textit{ConvNet}, inclusion of domain adaptation reduced the accuracy in the source domain by $1\%$, but allowed the model to perform with $9\%$ better accuracy in the target domain. It also increased the distance to successfully flipped one-pixel perturbed images by a factor of ${\approx}1.3$.

Domain adaptation methods can help bring discrepant data distributions closer together even if the differences between the datasets are quite large. 
Still, the best results are achieved when the datasets are preprocessed to be as similar as possible and include a large number of images for training. 
This is particularly important when one of the datasets contains simulated images, since one can work to make the simulations as realistic as possible, closer to the real data that that the model is intended to be used on.

Even though MMD has proven to be very successful in bridging the gap between astronomical datasets, it should not be used for very complex problems. 
MMD is a method that is not class-aware and hence tries to align entire data distributions. 
This property can be problematic when the two data distributions are very different from each other or when one of the datasets contains a new or unknown class that should not be aligned with the other domain. 
Our future work will focus on leveraging more sophisticated class-aware DA methods such as Contrastive Adaptation Networks~\citep[CAN;][]{KJ2019}, or Domain Adaptive Neighborhood Clustering via Entropy optimization~\citep[DANCE;][]{SK2020}, which can successfully perform domain alignment in more complex experiments. 
Note that even these more sophisticated methods work better if the similarity between datasets is greater or when the two datasets include more overlapping classes.

Although we adopted a generic \textit{ResNet18} to carry out our experiments with domain adaptation for robustness, other adversarial robustness approaches, such as those based on architecture improvement, data augmentation, and probabilistic modeling, can be used alongside domain adaptation. This is a future direction we will pursue.

In astronomy, new insights about astrophysical objects often come from our ability to simultaneously learn from multiple datasets: simulated and observed, observations from different telescopes and at different wavelengths, or using the same observations but with different observing times. 
Domain adaptation techniques are ideally suited in cases where deep learning models need to work in multiple domains and can even work when one of the domains is unlabeled. 

In many astrophysics applications, it is very difficult to precisely simulate the objects we are studying, and the observations themselves can be very noisy and include artifacts and detector errors. Our ability to build models that can overcome these difficulties is paramount, especially in cases where machine learning can be employed to help with rare or difficult to detect events. For example, machine learning has already been shown to help in detecting gravitational waves~\citep{GH2018,MO2022,AC2022}. Both precise simulations and dealing with very noisy observations makes detection challenging: gravitational wave detectors are very sensitive to noise from the physical environment, seismic activity, and complications in the detector itself (the data stream can contain sharp lines in its noise spectrum and non-Gaussian transients, or ``glitches''~\citep{CC2020}, that are not astrophysical in origin). Low-surface brightness galaxies are another type of difficult object that has been shown to be detectable by machine learning methods~\citep{TC2021}. When dealing with very faint objects, small data perturbations or other faint artifacts~\citep{TC2021b} can reduce our ability to find and characterize them in new survey data, so development of robust machine learning methods is important.

In scientific applications, where data perturbations are typically not targeted, but rather occur naturally, using domain adaptation can simultaneously help a) increase robustness to these small perturbations and b) realize the gains when information comes from multiple datasets. 
Future developments and implementations of adversarial robustness and domain adaptation methods in astronomical pipelines will open doors for many more uses of deep learning models.

\section*{Acknowledgments}

This manuscript has been supported by Fermi Research Alliance, LLC under Contract No.\ DE-AC02-07CH11359 with the U.S.\ Department of Energy (DOE), Office of Science, Office of High Energy Physics. This research has been partially supported by the High Velocity Artificial Intelligence grant as part of the DOE High Energy Physics Computational HEP program. 
This research has been partially supported by the DOE Office of Science, Office of Advanced Scientific Computing Research, applied mathematics and SciDAC programs under Contract No.\ DE-AC02-06CH11357. 
This research used resources of the Argonne Leadership Computing Facility at Argonne National Laboratory, which is a user facility supported by the DOE Office of Science.

The authors of this paper have committed themselves to performing this work in an equitable, inclusive, and just environment, and we hold ourselves accountable, believing that the best science is contingent on a good research environment.
We acknowledge the Deep Skies Lab as a community of multi-domain experts and collaborators who have facilitated an environment of open discussion, idea-generation, and collaboration. This community was important for the development of this project.

We are very thankful to Nic Ford for useful discussion regarding church window plots. Furthermore, we also thank the two anonymous referees who helped improve this manuscript.

\subsection*{\it{Author Contributions}}

A.~\'Ciprijanovi\'c: \textit{Conceptualization, Data curation, Formal analysis, Investigation, Methodology, Project administration, Resources, Software, Supervision, Visualization, Writing of original draft}; D.~Kafkes: \textit{Formal analysis, Investigation, Methodology, Resources, Software, Visualization, Writing of original draft}; S.~Madireddy: \textit{Conceptualization, Methodology, Resources, Software, Supervision, Writing (review \& editing)}; B.~Nord: \textit{Conceptualization, Methodology, Supervision, Writing (review \& editing)}; K.~Pedro: \textit{Conceptualization, Methodology, Project administration, Resources, Software, Supervision, Writing  (review \& editing)}; G.~N.~Perdue: \textit{Conceptualization, Methodology, Project administration, Resources, Software, Supervision, Writing  (review \& editing)}; F.~J.~S\'{a}nchez: \textit{Data curation, Methodology, Writing (review \& editing)}; G.~F.~Snyder: \textit{Conceptualization, Data curation, Methodology, Writing (review \& editing)}; S.~M.~Wild: \textit{Conceptualization, Methodology, Writing (review \& editing)}.

\section*{Data and Code Availability Statement}

The data that support the findings of this study are openly available at the following URL/DOI:
\href{https://doi.org/10.5281/zenodo.5514180}{https://doi.org/10.5281/zenodo.5514180}. 

The code that was used to perform the experiments presented in this paper is openly available in our GitHub repository:

\noindent\href{https://github.com/AleksCipri/DeepAdversaries}{https://github.com/AleksCipri/DeepAdversaries}.


\bibliographystyle{model2-names.bst}
\bibliography{main}

\end{document}